\documentclass{article}

\usepackage{microtype}
\usepackage{graphicx}
\usepackage{booktabs} 
\usepackage{tabularx}
\usepackage{caption}
\usepackage{float}
\usepackage{subcaption}

\usepackage{hyperref}

\usepackage[accepted]{icml2024}

\usepackage{amsmath}
\usepackage{amssymb}
\usepackage{mathtools}
\usepackage{amsthm}

\usepackage[capitalize,noabbrev]{cleveref}

\theoremstyle{plain}

\theoremstyle{definition}

\theoremstyle{remark}

\usepackage[textsize=tiny]{todonotes}

\begin{document}

\onecolumn
\icmltitle{PathoLM: Identifying pathogenicity from the DNA
sequence through the Genome Foundation Model}



\icmlsetsymbol{equal}{*}

\begin{icmlauthorlist}
\icmlauthor{Sajib Acharjee Dip}{cs}
\icmlauthor{Uddip Acharjee Shuvo}{iit}
\icmlauthor{Tran Chau}{gbcb}
\icmlauthor{Haoqiu Song}{cs}
\icmlauthor{Petra Choi}{cee}
\icmlauthor{Xuan Wang}{cs}
\icmlauthor{Liqing Zhang}{cs,gbcb}
\end{icmlauthorlist}

\icmlaffiliation{cs}{Department of Computer Science, Virginia Tech, USA}
\icmlaffiliation{gbcb}{Department of Genetics, Bioinformatics and Computational Biology program (GBCB), Virginia Tech, USA}
\icmlaffiliation{cee}{Department of Civil and Environmental Engineering, Virginia Tech, USA}
\icmlaffiliation{iit}{Institute of Information Technology, Dhaka University, Bangladesh}

\icmlcorrespondingauthor{Sajib Acharjee Dip}{sajibacharjeedip@vt.edu}
\icmlcorrespondingauthor{Liqing Zhang}{lqzhang@vt.edu}

\icmlkeywords{Machine Learning, ICML}

\vskip 0.3in



\printAffiliationsAndNotice 

\begin{abstract} 
Pathogen identification is pivotal in diagnosing, treating, and preventing diseases, crucial for controlling infections and safeguarding public health. Traditional alignment-based methods, though widely used, are computationally intense and reliant on extensive reference databases, often failing to detect novel pathogens due to their low sensitivity and specificity. Similarly, conventional machine learning techniques, while promising, require large annotated datasets and extensive feature engineering and are prone to overfitting. Addressing these challenges, we introduce PathoLM, a cutting-edge pathogen language model optimized for the identification of pathogenicity in bacterial and viral sequences. Leveraging the strengths of pre-trained DNA models such as the Nucleotide Transformer, PathoLM requires minimal data for fine-tuning, thereby enhancing pathogen detection capabilities. It effectively captures a broader genomic context, significantly improving the identification of novel and divergent pathogens. We developed a comprehensive data set comprising approximately 30 species of viruses and bacteria, including ESKAPEE pathogens, seven notably virulent bacterial strains resistant to antibiotics. Additionally, we curated a species classification dataset centered specifically on the ESKAPEE group. In comparative assessments, PathoLM dramatically outperforms existing models like DciPatho, demonstrating robust zero-shot and few-shot capabilities. Furthermore, we expanded PathoLM-Sp for ESKAPEE species classification, where it showed superior performance compared to other advanced deep learning methods, despite the complexities of the task. 
\end{abstract}

\section{Introduction}
Pathogens, ranging from flu viruses to severe disease agents like tuberculosis, can cause significant health challenges, leading to high morbidity and mortality, particularly in areas with limited healthcare. The emergence of antibiotic resistance further complicates treatment, elevating the risk of minor infections becoming fatal. Highlighted by the COVID-19 pandemic, pathogen surveillance using next-generation sequencing (NGS) is critical for monitoring public health across various settings. Effective pathogen identification is essential for timely disease management, informing treatment strategies, and fostering advancements in medical research.

Pathogens, including viruses, bacteria, fungi, and parasites, are major causes of infectious diseases worldwide. Their rapid mutation and reproduction make timely pathogen identification crucial for effective interventions. However, labeled data for these pathogens is scarce. To address this, we curated a dataset focused on ESKAPEE \cite{ruekit2022molecular} and viral pathogens, using pathogenic strains from the PATRIC database and non-pathogenic strains from NCBI. This strategic compilation provided a robust dataset allows the PathoLM model to effectively distinguish between pathogenic and non-pathogenic strains. To enhance PathoLM's pathogen detection, we expanded our dataset with additional viral sequences from NCBI, targeting human pathogens for the positive sample pool and using animal pathogens as negatives. This strategy broadened our training scope to include about 30 diverse species, creating a well-balanced dataset for our pathogen language model.

Previous pathogen detection methods used alignment-based techniques that struggle with novel pathogens and require intensive computation. To address these challenges, recent advancements have integrated machine learning and deep learning strategies, aiming to improve data classification and analysis. However, these advanced methods often require complex feature engineering and large well-annotated datasets for effective training. Many utilize k-mer frequency-based features, which despite their efficacy struggle with scalability and sensitivity, particularly in the detection of new pathogens. Specifically, k-mer-based techniques are adept at analyzing short reads but face obstacles with large-scale data integration and are prone to overfitting \cite{lorenzi2020imoka, watts2019identification, liebhoff2023pathogen}. As for our knowledge, the latest method, DciPatho, enhances pathogen identification by combining k-mer frequency features ranging from 3 to 7 k-mers and integrates three computational models—ResNet, DeepNet, and CrossNet—to create a robust intermediate representation. This approach significantly outperforms previous methods like PaPrBaG \cite{deneke2017paprbag}, BacPaCS \cite{barash2019bacpacs}, and DeePac \cite{bartoszewicz2020deepac}. Despite its improvements, DciPatho still faces limitations related to computational time for training from scratch, and its performance heavily depends on the quality and size of the training dataset.

Recent advancements in large foundation models have shown significant promise in various fields, including biomedicine and genomics \cite{li2021bioseq, luu2024bioinspiredllm, bi2024ai}. These models, trained on extensive datasets, capture complex patterns and relationships in data, which can be leveraged to improve pathogen detection. Utilizing pre-trained language models strategically mitigates the need for extensive domain-specific datasets and computational power. In addressing these challenges, we introduced PathoLM, a genome modeling tool that uses the pre-trained Nucleotide Transformer v2 50M \cite{dalla2023nucleotide} for enhanced pathogen detection in bacterial and viral genomes, both improving accuracy and addressing data limitations. To the best of our knowledge, we are the first to leverage the pre-trained knowledge of DNA language model for pathogen prediction tasks. Moreover, we have demonstrated PathoLM's capabilities using different foundation models as backbone, in zero-shot and few-shot scenarios, showcasing its robust performance compared to traditional machine learning and deep learning methods in species classification. This model not only addresses the challenges of limited data availability but also sets a new standard in pathogen detection technology.

\section{Materials and Methods
}\label{sec2}


\subsection{Data collection}

\paragraph{ESKAPEE genome} 
For the pathogen dataset, we downloaded 49,642 whole genome assemblies from the PATRIC website \cite {gillespie2011patric}, encompassing approximately 27 species of bacteria. From this collection, we retained only the genomes of the seven ESKAPEE pathogens: Escherichia coli, Enterococcus faecium, Staphylococcus aureus, Klebsiella pneumoniae, Acinetobacter baumannii, Pseudomonas aeruginosa, and all species from the Enterobacter genus. These genomes varied significantly in size, ranging from 200 to 1,147,640 base pairs. For the Eskapee non-pathogen data, we sourced our dataset from the NCBI database, ensuring to exclude any strains identified as pathogens in the PATRIC database. We obtained 10,633 whole genome sequences, which originally included around 22 different species.
\begin{table*}[h]
    \setlength{\tabcolsep}{5pt} 
    \renewcommand{\arraystretch}{1} 
    \centering
    \begin{small}
    \begin{sc}
    \caption{Viral data set for pathogenic and non-pathogenic species}
    \label{tab:my_label}
    \begin{tabularx}{\textwidth}{|X|X|X|X|}
        \hline
        Virus & Pathogenic & Nonpathogenic & References \\
        \hline
        Coronavirus & SARS, MERS, OC43, NL63, 229E, HKU1 & Bovine Coronavirus, Canine coronavirus, Feline coronavirus, Equine Coronavirus, Bat coronavirus, Mink coronavirus, Murine coronavirus, Avian coronavirus & \cite{saib2021non} \\
        \hline
        Influenza & type A, B, C & type D & \cite{cdc_influenza_website} \\
        \hline
        Norovirus & GI, GII, GIV & GIII, GV, GVI, GVII, GVIII & \cite{hassan2019norovirus} \\
        \hline
        Immunodeficiency virus & HIV & FIV, BIV, SIV & \cite{egberink1992animal} \\
        \hline
    \end{tabularx}
    \end{sc}
    \end{small}
\end{table*}

\paragraph{Viral genome} All viral genome data were downloaded from NCBI. For specific binary prediction, we limited our train dataset to include species that has both human pathogenic and non pathogenic virus. For coronavirus, SARS, MERS, OC43, NL63, 229E, HKU1 strains were considered as pathogenic \cite{saib2021non}. Bovine Coronavirus, Canine coronavirus, Feline coronavirus, Equine Coronavirus, Bat coronavirus, Mink coronavirus, Murine coronavirus, Avian coronavirus were considered as nonpathogenic. For influenza, type A, B, and C were considered human pathogenic while type D was considered as nonpathogenic \cite{cdc_influenza_website}. For norovirus, strain GI, GII, GIV were considered as human pathogenic strains while GIII, GV, GVI, GVII, GVIII were considered as non pathogenic virus  \cite{hassan2019norovirus}
\cite{egberink1992animal}. To address data disparities in health surveillance, where non-pathogenic data are less abundant than human pathogenic data due to a focus on clinical patients, we expanded our dataset. We included non-pathogenic viruses often found in wastewater metagenomics, such as Escherichia phage T1 \& T7, Bacteriophage sp., Klebsiella pneumoniae phage, and Crassphage, sourced from NCBI. Additionally, we gathered genomes of plant pathogens like Tomato yellow leaf curl virus, pepper mild mottle virus, and tomato mosaic virus to diversify our dataset further.

\subsection{Data Processing}
\subsubsection{Data Filtering and Integration}
Our data preparation process began with the cleaning and organization of ESKAPEE pathogen and non-pathogen datasets separately. We simplified this data set to include only seven species, mirroring the selection criteria used for the pathogen data. The sequence lengths in this dataset were more varied, ranging from 294 to 7,411,863 base pairs. We expanded our binary prediction model by incorporating viral genomes alongside bacterial ones. The viral dataset included 25,229 complete genomes across 765 strains from species with at least 10 genomes each, such as Influenza A and B, Norovirus, HIV, and SARS-CoV-2, among others. Species names were standardized, and for nonpathogenic viruses, we analyzed 1,782 genomes, focusing on the top 14 species. We then integrated these with the ESKAPEE bacterial data, creating a comprehensive dataset for our binary prediction model. Training and testing sets were organized using MMseqs2 clustering to capture the genetic diversity, with detailed clustering methods outlined later.

\subsubsection{Data partitioning using sequence clustering }
To ensure the creation of a robust and generalizable evaluation dataset, we employed MMseqs2 \cite{steinegger2017mmseqs2} for clustering based on sequence similarity. This approach allowed us to systematically explore the impact of similarity thresholds on model performance. Specifically, we establish clusters using coverage and similarity thresholds set at 80\%, 60\%, and 40\%, resulting in three distinct datasets for subsequent experimentation. By selecting the highest sensitivity settings for the clustering process, we aimed to capture the most detailed relationships within our data. We subsequently allocated the generated clusters into training and test sets using an 8:2 split, with the assurance that both sets comprised different clusters. This methodical division was designed to foster a well-generalized dataset, leveraging the advanced capabilities of the MMseqs2 clustering suite to ensure that our train-test splits reflected true biological variations and complexities.

\subsubsection{Creating Varied Sequence Lengths for Performance Sensitivity Analysis}
To evaluate the impact of sequence length on model performance, we conducted experiments using a range of sequence lengths derived from whole genome assemblies. Specifically, we fragmented these assemblies into segments of varying lengths, including 150, 500, 2k, 5k, 10k, 50k, as well as using the entire genome sequence as input. For each specified length, we generated a separate dataset. To ensure the integrity of our experimental data, we applied MMseqs2 \cite{steinegger2017mmseqs2} clustering to each dataset to eliminate highly similar sequences, utilizing both coverage and similarity thresholds. This step was crucial for maintaining diversity within our data and ensuring that our findings were not biased by over represented sequences.


\subsection{Pretrained model}
\subsubsection{Pretraining dataset}
\begin{figure}[htbp]
    \centering
    \includegraphics[width=0.8\textwidth]{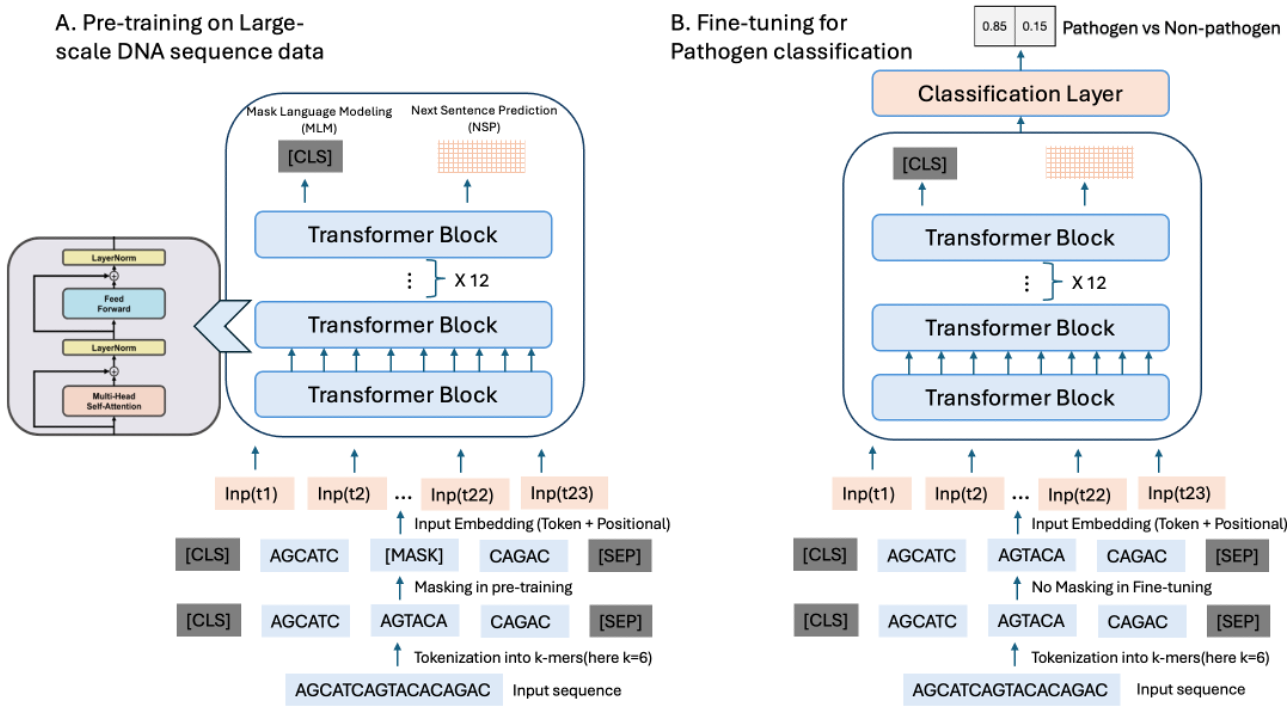}
    \caption{Model Architecture of PathoLM. (A) Pretraining steps of the genome foundation model on a large-scale dataset, where tokenization and masking are applied before feeding the input sequence into the transformer block. (B) Finetuning steps for the pathogen identification task using specific datasets, where no masking is applied.}
    \label{fig:pathoLM-archi}
\end{figure}

We utilized the nucleotide transformer multispecies v2 (50 million parameter) model  \cite{dalla2023nucleotide}. This model, pre-trained on a dataset including 3,202 human genomes and 850 genomes from diverse taxa but excluding plants and viruses, shown in Figure \ref{fig:pathoLM-archi}A, was chosen for its balance of performance and computational efficiency compared to its 250 million parameter counterpart. The curated dataset from NCBI featured one species per genus to ensure wide phylogenetic representation, totaling 174 billion nucleotides. It spans several taxa, including key organisms like Escherichia coli, Saccharomyces cerevisiae, Caenorhabditis elegans, various Plasmodium species, and vertebrates like Homo sapiens and Mus musculus.


\subsubsection{k-mer based tokenization}
In natural language processing, we split sentences into words using grammar and semantics, but genomic sequences lack clear delimiters, making tokenization challenging. A typical method is k-mer tokenization, where sequences are divided into overlapping substrings of a fixed length. In this study, we used a specialized tokenizer for nucleic acids, capable of recognizing 4,096 unique 6-mer combinations plus the nucleotides A, T, C, G, and N. This tokenizer also includes special tokens like `[PAD]`, `[MASK]`, and `[CLS]` for padding, masking, and classification, resulting in a total vocabulary of 4,104 tokens. The process begins with a `[CLS]` token and progresses by converting sequences into tokens from left to right, giving priority to 6-mer tokens and defaulting to individual nucleotide tokens where necessary. This method preserves biological information for deep learning applications, enhancing the pretrained Nucleotide Transformer model's accuracy in pathogen prediction by fine-tuning it on these tokenized sequences.

\subsubsection{Basic architecture of Nucleotide transformer}
The Nucleotide Transformer v2 50M, designed for genomic data, utilizes an encoder-only transformer architecture. It starts by embedding 6-mer sequences into dense vectors, enhanced with Rotary Positional Embeddings for positional context up to 12kb. The model employs multi-head self-attention within each transformer layer to analyze sequence interdependencies, incorporating residual connections to retain information. After normalization, the sequences proceed through a feed-forward network using Gated Linear Units and Swish activation to boost processing efficiency. In training, it predicts nucleotide occurrences to improve genomic data interpretation.

\subsection{Fine-tuning}
We fine-tuned our pre-trained Nucleotide Transformer model using the Hugging Face Transformers library, selected for its robust support for genomic sequences. We adapted the model to manage different sequence lengths, padding shorter sequences and truncating longer ones beyond the 12,000 nucleotide context limit shown in Figure \ref{fig:pathoLM-archi}B. The model was optimized for both binary classification of pathogens and detailed classification among ESKAPEE species, using dual and septenary output labels, respectively. Training involved the Adam optimizer with a specific learning rate and a warm-up phase to prevent premature convergence, coupled with an early stopping mechanism to avoid overfitting. This strategy effectively leveraged the Nucleotide Transformer's strengths, yielding precise and robust models for pathogen genomics.
\subsection{Evaluation}\label{subsec5}
We evaluated our model using key metrics: Accuracy, F1-Score, MCC, and AUC-ROC. Accuracy measures the rate of correct predictions, F1-Score assesses precision and recall balance, MCC provides a robust correlation coefficient regardless of class size, and AUC-ROC evaluates the model's ability to distinguish between classes at various thresholds. These metrics collectively ensure a thorough assessment of precision, recall, classification quality, and discriminative capacity.

\[
\text{Accuracy} = \frac{TP + TN}{TP + TN + FP + FN}
\]
Where \( TP \) = True Positives, \( TN \) = True Negatives, \( FP \) = False Positives, \( FN \) = False Negatives.

\[
\text{F1-Score} = 2 \times \frac{\text{Precision} \times \text{Recall}}{\text{Precision} + \text{Recall}}
\]
Where \( \text{Precision} = \frac{TP}{TP + FP} \) and \( \text{Recall} = \frac{TP}{TP + FN} \).

\[
\text{MCC} = \frac{TP \times TN - FP \times FN}{\sqrt{(TP+FP)(TP+FN)(TN+FP)(TN+FN)}}
\]



\section{Results}\label{sec3}

 \begin{figure}[ht]
    \begin{center}
    \centerline{\includegraphics[width=0.9\textwidth]{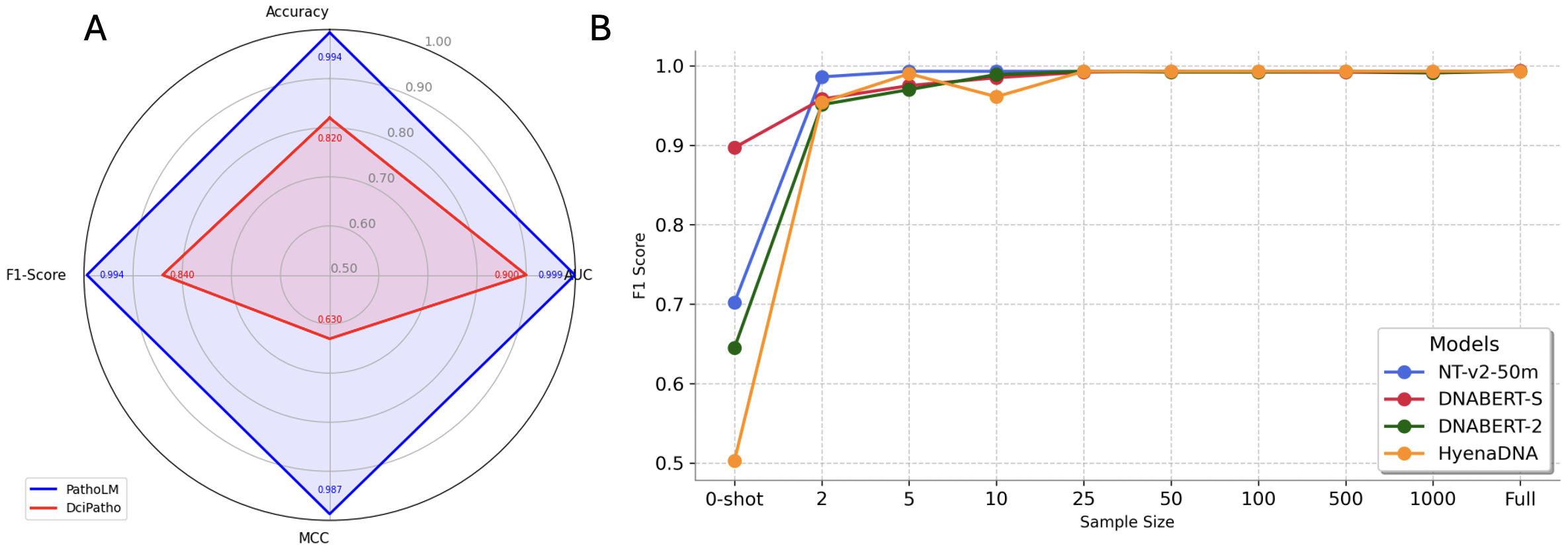}}
    \caption{A: Performance comparison between PathoLM and state-of-the-art method DciPatho on the binary pathogen prediction task in terms of Accuracy, F1-score, AUC-ROC, MCC. B: Comparison of F1-scores for various genome language models, highlighting their performance in zero-shot and few-shot scenarios.}
    \label{fig:patholm}
    \end{center}
\end{figure}
\subsection{Identifying pathogenicity from PATRIC ESKAPEE species and human pathogen virus}
The PathoLM model is built upon the pre-trained foundation model, Nucleotide Transformer. To adapt the pre-trained model parameters for our specific pathogen identification task, we employed transfer learning techniques. Specifically, we used the frozen weights from the Nucleotide Transformer and fine-tuned them on our pathogen identification dataset. To assess the effectiveness of PathoLM, we conducted a comparison with state-of-the-art method, including DciPatho \cite{jiang2023dcipatho} on a binary prediction task. Both PathoLM and DciPatho were trained on the same training dataset and evaluated on the same held-out test dataset to ensure a fair and consistent evaluation. As shown in Figure \ref{fig:patholm}A, PathoLM significantly outperforms DciPatho across all evaluation metrics, thanks to the pre-trained Nucleotide Transformer. This foundation enhances PathoLM's ability to detect subtle genomic patterns essential for accurate pathogen identification. Leveraging pre-trained representations allows PathoLM to deliver high accuracy and reliability in pathogen detection, highlighting the effectiveness of transfer learning in genomic applications.



\subsection{Zero-Shot and Few-Shot Capabilities in Large Language Models for Pathogenicity Identification}
While traditional machine learning and deep learning models require supervised training with labeled data, large language foundation models facilitate zero-shot classification, classing sequences without labeled data. We compared the zero-shot pathogen classification performance of four different language models pre-trained in a self-supervised manner on large-scale DNA sequence data, each differing in dataset and training methodologies. For instance, DNABERT and Nucleotide Transformer primarily utilize k-mer tokenization, while DNABERT-2 \cite{zhou2023dnabert} employs byte-pair encoding. DNABERT-S \cite{zhou2024dnabert} boosts performance through Manifold Instance Mixup and Curriculum Contrastive Learning. Our findings indicate that DNABERT-S excels in zero-shot tasks with a 90\% F1-score, while NT-v2-50m achieves a commendable 70\%, likely due to its extensive multi-species training. DNABERT-2 and HyenaDNA lag with performances less than 70\% shown in Figure \ref{fig:patholm}B. Additionally, we assessed the few-shot capabilities of these models by training with incrementally larger samples, observing optimal performance at 25 samples, showcasing the strong few-shot learning potential of large language models. Despite NT-v2-50m's lower initial zero-shot performance, it rapidly improves with minimal data input, surpassing others with only two samples from pathogen and non-pathogen categories, making it our model of choice due to its broad training on multiple species and lower computational demands compared to DNABERT-S.

\begin{figure}[ht]
    \vskip 0.1in
    \begin{center}    \centerline{\includegraphics[width=1\textwidth]{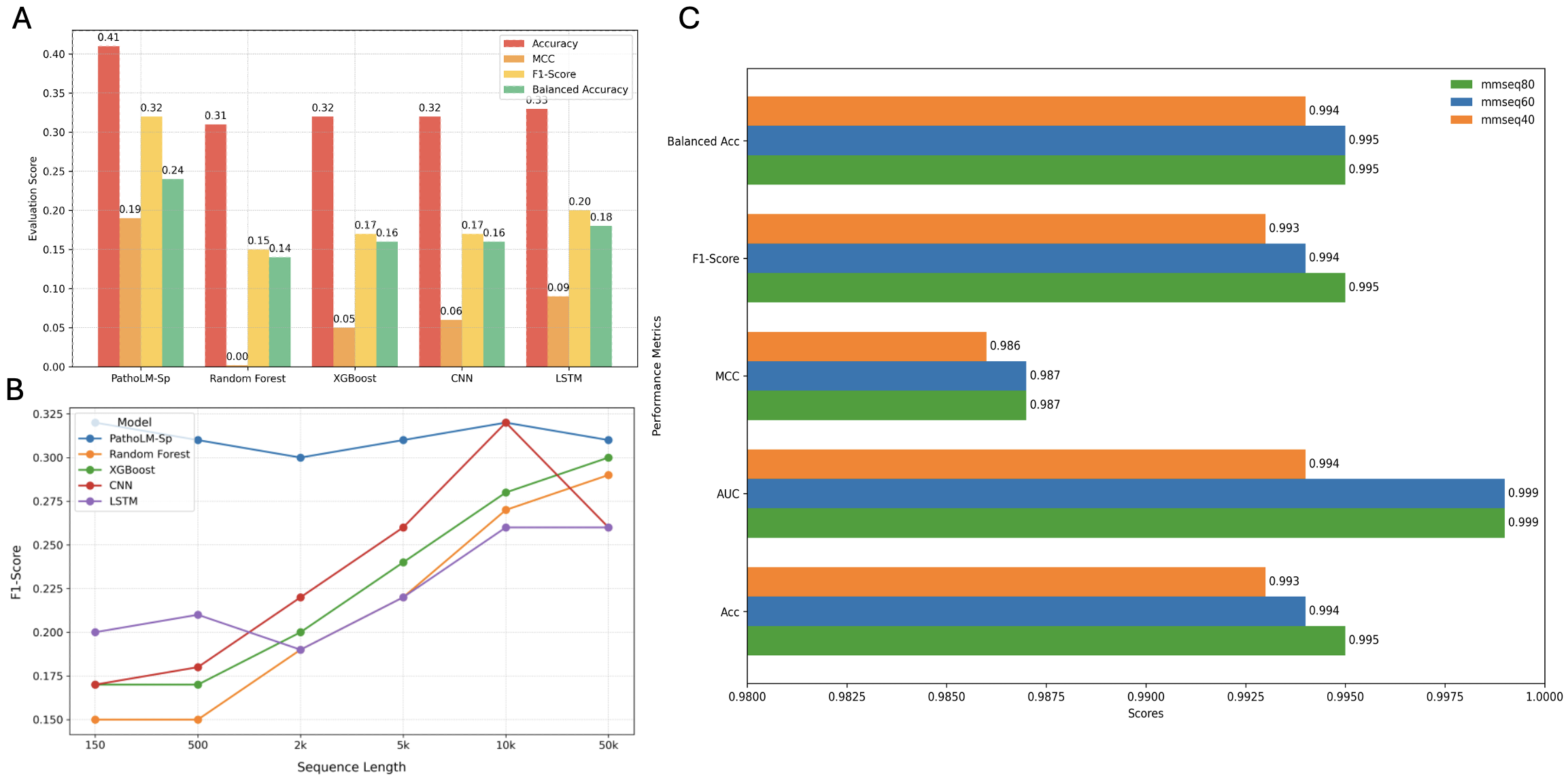}}
    \caption{A: Performance comparison between PathoLM-Sp and machine learning Random Forest, XGBoost and deep learning methods CNN, LSTM on the ESKAPEE species classification task. PathoLM-Sp outperforms all other methods across multiple evaluation metrics. B: Performance comparison of PathoLM-Sp and other methods for varied sequence length from 150 bp to 50k bp. PathoLM-Sp consistently performs well for all sequence lengths. C: Performance comparison of PathoLM for different train and test set data created using mmseq2 sequence clustering. The similiarity-threshold and coverage 80, 60, 40 are shown as mmseq80, mmse60, mmseq40.}
    \label{fig:patholm-sp}
    \end{center}
    \vskip -0.1in
\end{figure}
\subsection{Species classification with PathoLM-Sp}
We expanded our PathoLM model to classify seven ESKAPEE species, naming this variant PathoLM-Sp. Since there were no pre-trained models for ESKAPEE species, we compared PathoLM-Sp with machine learning methods like Random Forest and XGBoost, and deep learning methods including a 1-layer CNN and a 1-layer LSTM. Sequences were transformed into one-hot encodings and utilized as input features for the model. The CNN had a one-dimensional convolutional layer with 64 filters, ReLU activation, and max pooling, followed by two dense layers. The LSTM model used a single layer with 50 units. Both were optimized with Adam and trained on the same dataset for consistent evaluation. PathoLM-Sp demonstrated superior accuracy, F1-score, MCC, and balanced accuracy, as shown in Figure \ref{fig:patholm-sp}A, confirming its effectiveness despite the inherent challenge of species classification with these data.

\subsection{Performance comparison for varied sequence length
}
We evaluated PathoLM-Sp against other models using datasets of various sequence lengths (150 bp to 50k bp) to determine how sequence length impacts performance. PathoLM-Sp consistently outperformed all models at each sequence length, maintaining superior accuracy and robustness. While models like Random Forest, XGBoost, CNN, and LSTM improved with longer sequences, PathoLM-Sp excelled even with shorter sequences, leveraging its pre-trained capabilities effectively shown in Figure \ref{fig:patholm-sp}B. This performance consistency across various sequence lengths highlights PathoLM-Sp's reliability for genomic analysis, particularly when dealing with diverse data conditions.


\subsection{Performance Comparison on Different Train-Test Sets Created Using MMseq2}

To ensure a robust and generalized evaluation, we used MMseq2 to create different train-test splits based on sequence clustering. MMseq2 was employed to generate clusters using similarity threshold and coverage criteria. The clusters were divided into training and testing sets at an 8:2 ratio, ensuring samples from different clusters for each set. We created three distinct datasets with similarity thresholds and coverage values of 80, 60, and 40. The performance of our PathoLM-Sp model was compared against other models using these datasets shown in Figure \ref{fig:patholm-sp}C. The results demonstrate that model performance improves as the similarity threshold increases. Specifically, with an 80\% similarity threshold, the model achieves higher evaluation metrics compared to the 60\% and 40\% thresholds. This comparison highlights the impact of sequence similarity on model performance. A reduced similarity threshold enhances learning robustness and generalization, as evidenced by the variation in performance metrics.

\section{Discussion}\label{sec4}
In this study, we introduce PathoLM, a novel language model tailored for pathogen identification, drawing significantly from a pre-trained foundational model and a meticulously curated dataset of high-risk pathogens. This foundation imbues PathoLM with a deep genomic understanding, enabling it to discern subtle differences between pathogenic and nonpathogenic sequences across viruses and ESKAPEE bacteria, thereby enhancing its performance. Our database, concentrated on about 30 bacterial and viral species with a focus on ESKAPEE pathogens, supports extensive pathogen prediction tasks and offers a valuable resource for further research. Additionally, our PathoLM-Sp variant, designed for ESKAPEE species classification, surpasses conventional machine and deep learning methods in accuracy, highlighting the effectiveness and flexibility of our model.

Despite these advances, our model has limitations, notably the Nucleotide Transformer v2 50M's maximum context length of 12k bp. Sequences exceeding this length must be fragmented for analysis, potentially missing broader sequence interactions. Moreover, training high-parameter language models demands extensive computational resources. While the 50 million parameter model is manageable, the more accurate 250 million parameter version is even more resource-intensive. Additionally, PathoLM-Sp’s species classification accuracy around 41\% suggests room for improvement, particularly for species beyond E. coli. Future work will explore integrating diverse features to enhance model performance. Despite these challenges, our work significantly advances pathogen identification using pre-trained language models and provides a valuable dataset for ESKAPEE species, underscoring the potential for further enhancements and the current effectiveness of our approach.

\section{Code and Data Availability}
 We downloaded the ESKAPEE dataset from PATRIC website [\href{https://www.bv-brc.org/}{https://www.bv-brc.org/}] and downloaded viral and nonpathogenic ESKAPEE data from NCBI website [\href{https://www.ncbi.nlm.nih.gov/}{https://www.ncbi.nlm.nih.gov/}]. The source code and data of PathoLM will be revealed after/during review process.


\bibliography{example_paper}
\bibliographystyle{icml2024}




\end{document}